\documentclass[11pt,a4paper]{article}
\usepackage[hyperref]{acl2020}
\usepackage{times}
\usepackage{latexsym}

\usepackage{microtype}
\usepackage{graphicx}
\usepackage{lipsum}
\graphicspath{ {./} }
\aclfinalcopy

\title{EmotionGIF-Yankee: A Sentiment Classifier with Robust Model Based Ensemble Methods}

\author{Wei-Yao Wang\textsuperscript{1*}, Kai-Shiang Chang\textsuperscript{1} and Yu-Chien Tang\textsuperscript{2} \\
  \textsuperscript{1}Advanced Database System Laboratory, National Chiao Tung University, Hsinchu, Taiwan \\
  \textsuperscript{2}Department of Computer Science, National Chiao Tung University, Hsinchu, Taiwan\\
  \{sf1638.cs05, ksjhang60523.cs05, tommytyc.cs06\}@nctu.edu.tw}

\date{2020-07-04}

\begin{document}
\maketitle
\begin{abstract}
This paper provides a method to classify sentiment with robust model based ensemble methods.
We preprocess tweet data to enhance coverage of tokenizer.
To reduce domain bias, we first train tweet dataset for pre-trained language model.
Besides, each classifier has its strengths and weakness, we leverage different types of models with ensemble methods: average and power weighted sum.
From the experiments, we show that our approach has achieved positive effect for sentiment classification.
Our system reached third place among 26 teams from the evaluation in SocialNLP 2020 EmotionGIF competition.
\end{abstract}

\section{Introduction}
Natural language is often indicative of one's emotion. 
Hence, detecting emotions in textual conversations has been one of popular topics in the field of natural language processing (NLP) sentiment domain. 
Sentiment classifier can help researchers study such information on user's feeling. 
There are various tasks of sentiment classification, for example, \citet{riloff2005exploiting} presents an information extraction (IE) system that automatically uses filtering extractions to improve subjectivity classification. 
On opinion extraction, \citet{zhai2011exploiting} extracts different opinion feature, including sentiment-words, substrings, and key-substring-groups, to help improve sentiment classification performance. 
In recent years, \citet{hazarika2018icon} proposes a multi-modal emotion detection framework, interactive conversational memory network (ICON), to extract multi-modal features for emotion detection.

In SocialNLP 2020 EmotionGIF, the challenge is to use tweet text and reply to recommend exactly 6 categories. 
In this paper, we propose an architecture to apply to the shared task. 
We preprocess original tweet data to pre-trained language model, then fine-tune to multi-label classification model. 
To build comprehensive emotion classifier, we design an ensemble scheme to get higher performance.

\section{Dataset}
\label{2}
The shared task includes a first-of-its-kind dataset of 40,000 two-turn Twitter threads.
Each thread contains 5 columns which are \textbf{idx}, \textbf{text}, \textbf{reply}, \textbf{categories}, and \textbf{mp4}.

Here are the explanations of 5 columns:
\begin{itemize}
\item idx: a unique identifier of each tweet
\item text: the text of the original tweet
\item reply: the text content of the response tweet
\item categories: the categories of the response GIF, containing 1 to 6 categories out of a list of 43 categories
\item mp4: the hash file name of the response GIF
\end{itemize}

The dataset is split into three JSON files, \textbf{train-gold}, \textbf{dev-unlabeled}, and \textbf{test-unlabeled}.
First including 32,000 threads is training data, and the others including 4,000 threads are validation data, and testing data.
The difference between \textbf{train-gold} and \textbf{dev-unlabeled}, \textbf{test-unlabeled} is that the former consists of all the 5 columns while the latter two only consist of 3 columns, \textbf{idx}, \textbf{text}, and \textbf{reply}.

\begin{figure}
\includegraphics[width=\linewidth]{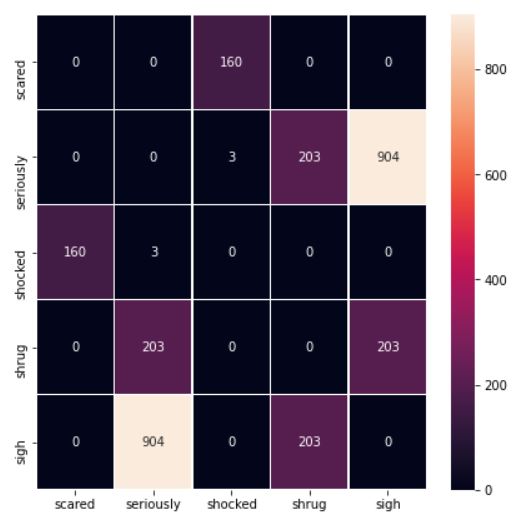}
\caption{Subset of correlation table}
\label{correlation table}
\end{figure}

\begin{figure*}
\includegraphics[width=\textwidth]{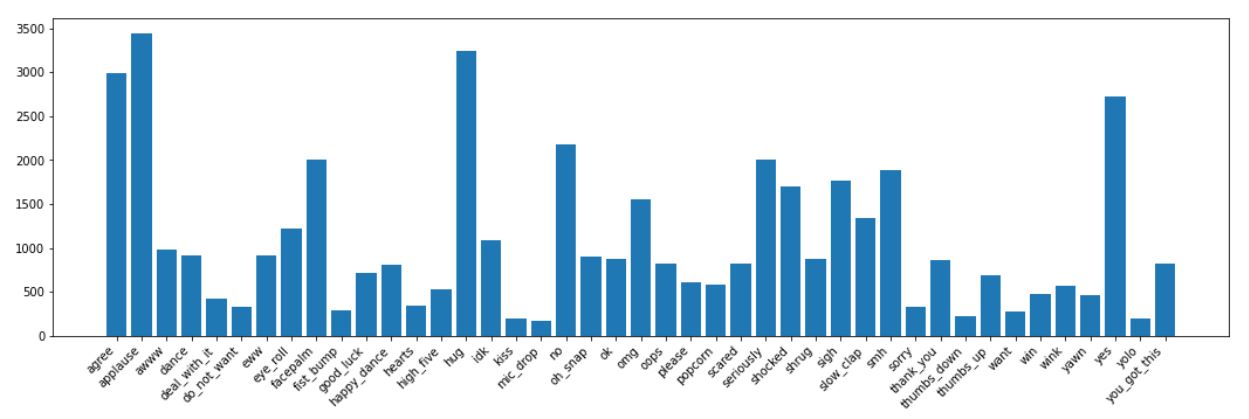}
\caption{Distribution of 43 categories}
\label{distribution}
\end{figure*}

Figure \ref{correlation table} is the subset of the correlation table which contains the frequency of co-appearance of any two categories.
The figure illustrates the correlation between different categories and we can observe that some categories have strong connection while some categories have weak connection.
Figure \ref{distribution} is the distribution of 43 categories in training data.
The figure shows that there is an imbalance between categories. 

\section{Related Work}
Our study can be mainly divided into three topics, including multi-label classification, pre-trained models, and ensemble methods.

\subsection{Multi-label Classification}
Multi-label classification is a generalization of multiclass classification. Nowadays the multi-label classification is increasingly used in many fields of NLP, such as semantic scene classification and sentiment classification.
There are two main categories of multi-label classification approaches: problem transformation (PT) methods, and algorithm adaptation (AA) methods.
Generally speaking, problem transformation methods will transform the multi-label classification problem into one or more single-label classification problem  \citep{zhang2014learingalgorithm}, while algorithm adaptation methods usually use those algorithms having been adapted to multi-label task and needing no problem transformation. 
For problem transformation methods, a good strategy to achieve the goal is Classifier Chains (CC) \citep{read2011classifier}, which classifies whether the original multi-label problem belongs to a label or not in chain structure, and is able to capture the interdependencies between the labels.
And for algorithm adaptation methods, Multi-label k-Nearest Neighbors (MLkNN) \citep{zhang2007ml} is one of the most popular.
It relies on the maximum a posteriori (MAP) principle on training the k-Nearest Neighbors (kNN), which is a well known traditional machine learning algorithm, to determine which label subset each instance belongs to. 
Due to its promising results and simplicity, it has been applied to many practical tasks of text classification.

Most of the existing multi-label classification approaches solve the emotion classification by training the model on a large dataset. 
The idea is to find informative features which can reflect the emotion expressed in the text, so with this approach most studies aim to find efficient features leading to better performance \citep{jabreel2016sentirich}.
Also, deep learning models are introduced to solve the multi-label classification problem, and have been proved that such models are able to extract high-level features from raw data.
For instance, \citet{baziotis2018ntua}, the winner of SemEval-2018 Task 1
competition: “Affect in Tweets”, proposes a Bi-LSTM architecture with attention mechanism.
They leverage a set of word2vec word embeddings trained on a dataset of 550 million tweets.

\subsection{Pre-trained Models}
Pre-trained models have been widely applied in a variety of NLP systems and achieve dramatically performance for downstream tasks.
There are three major advantages for pre-trained models.
First of all, since they are unsupervised learning, there will be unlimited corpus can be trained.
Secondly, a strength pre-trained language model can generate deep contextual word representation which means a word token can have several representation in different sentences.
Hence, through fine-tuning we improve downstream tasks more efficiently.
Last but not least, using pre-trained models can reduce huge architecture engineering.
This allows us don't need to design a deep learning network by ourselves and pre-train with massive cost.

BERT \citep{devlin2018bert}, Bidirectional Encoder Representations from Transformers, is one of state-of-the-art (SOTA) pre-trained model.
There are two main tasks in pre-training stages.
At the first task, called Masked LM (MLM), is to replace 15\% of the words in each sequence to a \texttt{[MASK]} token and model need to predict these masked tokens.
Encoder learns contextual representations during this stage.
Second task, Next Sentence Prediction (NSP), the model takes pairs of sentences as input and learns to predict if the second sentence in the pair is the subsequent sentence in the original documents. In details, 50\% of the inputs will be a pair in original documents in training, while the other 50\% a random sentence from the corpus is chosen as the second sentence.

There are some variant models based on BERT like RoBERTa \citep{liu2019roberta} and DistilBERT \citep{sanh2019distilbert}. 
DistilBERT, distilled BERT, reduces the size of a BERT model by 40\%, while retaining 97\% of its language understanding and being 60\% faster.
DistilBERT removes half number of layers on \textit{token-type embeddings} and the \textit{pooler}.
Instead of focusing on efficiency, RoBERTa, robustly optimized BERT approach, finds BERT undertrained that is why they study carefully to modify key hyperparameters to improve performance.
Since there is different discrepancy about whether to remove NSP \citep{devlin2018bert, lample2019cross, yang2019xlnet, joshi2020spanbert}, RoBERTa do some experiments and find that remove NSP can slighlty improve downstream tasks.
Furthermore, RoBERTa uses bytes instead of unicode as the base subword units \citep{radford2019language}.
Using bytes makes model learn larger subword vocabulary.

\subsection{Ensemble Methods}
In general, supervised learning can be defined as finding hypotheses (classifier) that are closed to the true function which can represent all the data points in training data.
However, learning algorithms that only output one hypothesis would face three major problems, \textbf{statistical}, \textbf{computational}, and \textbf{representational}.
Fortunately, ensemble methods construct a set of classifiers and then classify new data points by taking a vote of their predictions which could usually address the three problems just mentioned \citep{dietterich2000ensemble}.
The first problem is that learning algorithms may give same accuracy with different hypotheses.
By constructing an ensemble out of all of these accurate classifiers, the algorithm can use a simple and fair voting mechanism to reduce the risk of choosing the wrong classifier. 
The second problem is that many learning algorithms implement local search which may stop even if the best solution found by the algorithm is not optimal.
An ensemble constructed by running the local search from many different starting points may provide a better approximation to the true unknown function than any of the individual classifiers.
The third problem is that in most applications of machine learning, the true function cannot be represented by any of the hypotheses.
By forming weighted sums of hypotheses, it may be possible to expand the space of representable functions.

\citet{hagen2015webis} introduces an approach with ensemble methods on twitter sentiment detection. Their ensemble method is a voting scheme on the actual classifications of the individual classifiers rather than averaging confidences.
Their system proves a strong baseline in the SemEval 2015 evaluation.

These studies motivate us to transfer EmotionGIF task into multi-label classification problem since this task needs to infer most possible 6 categories of each tweet. 
To reduce huge architecture engineering, we adopt pre-trained models then focus on preprocessing and postprocessing stages such as ensemble methods to achieve better performance on the competition.
In addition to solving those problems mentioned previously, each classifier has its strengths and weakness, if we can combine different types of classifiers to leverage others forte to cover its own drawbacks, we can obtain highly accurate classifiers by combining less accurate ones.
By combining these three techniques, we could build a robust system on EmotionGIF task.

\section{Methodology}
\begin{figure*}
\centering
\includegraphics[width=\linewidth]{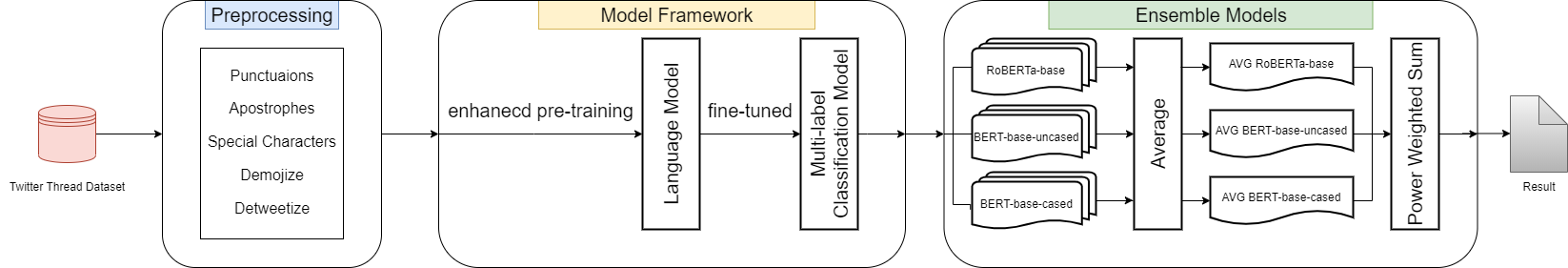}
\caption{Architecture of the system}
\label{architecture}
\end{figure*}

The main goal of the present work is to predict 6 most possible categories for each tweet in EmotionGIF task.
We propose an architecture as in Figure \ref{architecture} which includes three stages: preprocessing, model framework, and ensemble methods.

\subsection{Preprocessing}
Tweet data don't have same structure as formal corpus (e.g. Wikipedia).
There are multiple methods to clean up original tweet data.
We perform some methods to normalize data, including five steps, but we do not convert to lower case.
Here are the main five steps to normalize tweet dataset.
We do these steps in order:
\begin{enumerate}
\item Transform weird punctuation such as \textit{’} and \textit{‘}.
\item Transform apostrophes to original words. For example, \textit{hasn't} will be converted to \textit{has not}.
\item Mapping unknown punctuation which not in tokenizer's vocabulary. 
For example, \begin{math} \beta \end{math} is unknown in RoBERTa tokenizer, this will be transformed to word \textit{beta}.
\item Demojize: convert emoji symbols into their corresponding meanings. 
Also, if there are duplicate emojis, we will only retain one emoji to represent these duplicate emojis.
\item Detweetize and more words conversion: some words in dataset are in tweet style, which means these words are seldom seen in formal corpus. We replace these words by manually into common representations. 
Like \textit{idk} will be replaced with \textit{I don't know}.
Moreover, there are many recent trends like \textit{COVID} which haven't been seen in tokenizers before. Therefore, we transform these words to common words like \textit{virus} which can be tokenized correctly in tokenizers.
\end{enumerate}

\subsection{Model Framework}
Model framework is composed of two parts: enhanced pre-trained language model and fine-tuned multi-label classification model.

Pre-trained model trains on formal corpus like Wikipedia instead of tweet dataset. 
To avoid our model overfitting and domain bias on the training data, we use provided 32,000 training set to further train on pre-trained language model.
The enhanced language model understands more about tweet style sentences.

In EmotionGIF task, we treat as multi-label classification problem.
Hence, we use enhanced pre-trained model to fine-tune to multi-label classification model in downstream task.

To properly handle multi-label classification, we select \textit{BCEWithLogitsLoss} as our loss function. \textit{BCEWithLogitsLoss} combines a sigmoid layer and the BCELoss, and takes advantage of the log-sum-exp trick for numerical stability as Equation (\ref{BCEWithLogitsLoss}) and Equation (\ref{BCEWithLogitsLoss-detail}).
\begin{equation}
    l(x, y) = L = \{l_1, ..., l_N\}^T
\label{BCEWithLogitsLoss}
\end{equation}
where N is the batch size, 
\begin{equation}
    l_n = -w_n [y_n * log(x_n) + (1-y_n) * log(1-x_n)]
\label{BCEWithLogitsLoss-detail}
\end{equation}

Our goal aims to get better performance instead of efficiency, we use \textbf{RoBERTa-base}, \textbf{BERT-base-cased}, and \textbf{BERT-base-uncased} to individually train language model and fine-tune to multi-label classification model. 
Since RoBERTa and BERT use different input formats, and our dataset has pair of sequences \texttt{text} and \texttt{reply} in each tweet, we convert input sentences based on corresponding models.
BERT format is to add a special token \texttt{[CLS]} at first and add \texttt{[SEP]} between sentences and the end.
RoBERTa format is to add \textless s\textgreater \space at first and add \textless /s\textgreater \space between sentences and the end.
An example of representation is as Table \ref{representation-example}.

\begin{table*}
\centering
\begin{tabular}{llll}
\hline \textbf{Model} & \textbf{Text} & \textbf{Reply} & \textbf{Representation} \\ \hline
BERT & Don't forget to Hydrate! & & [CLS] Don't forget to Hydrate! [SEP]  [SEP]\\
RoBERTa & Don't forget to Hydrate! & & \textless s\textgreater \space Don't forget to Hydrate! \textless /s\textgreater\textless /s\textgreater \space \textless s\textgreater\\
\hline
\end{tabular}
\caption{\label{representation-example} An example of representation }
\end{table*}

\subsection{Ensemble Methods}

\begin{figure}
\centering
\includegraphics[width=\linewidth]{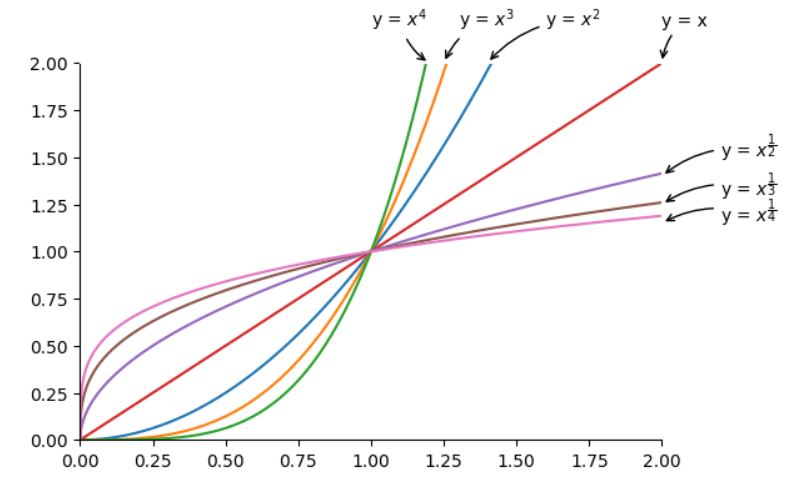}
\caption{Visualization of equation y = x\textsubscript{n} with n = 1/4, 1/3, 1/2, 1, 2, 3, 4}
\label{exponent}
\end{figure}
Since each classifier has its strengths and weakness, if we can combine different types of classifiers to leverage others forte to cover its own drawbacks, we can obtain highly accurate classifiers by combining less accurate ones.
To attain the desired results, we combine three different types of models, \textbf{RoBERTa-base}, \textbf{BERT-base-cased}, and \textbf{BERT-base-uncased}.
On account of different dropout weights in each training, the performance of each trained model may have a big gap compared with each others.
By training 10 same type of models with different dropout weights and averaging their predictions, we can lower the risk of using single model with bad performance.
\begin{equation}
    P_1 ^ N * w_1 + P_2 ^ N * w_2 + P_3 ^ N * w_3
\label{power weighted sum}
\end{equation}
After training and averaging three types of model, we use Equation (\ref{power weighted sum}) to get our final result where P\textsubscript{i} for i = 1, 2, 3 are average predict scores from \textbf{RoBERTa-base}, \textbf{BERT-base-cased} and \textbf{BERT-base-uncased} respectively and w\textsubscript{i} for i = 1, 2, 3 are the weights corresponding to each model.
To choose a reasonable N, we look into the property of power function. 
Figure \ref{exponent} shows that the further away the probability is from 1, then the faster the probability is closer to 0 and vice versa.
The probabilities that remain the highest at the end are the probabilities whose relative agreement (weighted down by the probability and the power) coming from each ensemble model is the highest \citep{laurae2020ensemble}.
We take advantage of the power weighted sum to enhance performance of model.

\section{Experiment}
In EmotionGIF, we only have ground truth labels in training data.
We use dev-unlabeled as our validation data.
That is, we fine-tune hyperparameters based on validation data and use best models from tuning to predict testing data, test-unlabeled.
In this section, our system gives some reasonable results from experiments.
The source code for this paper is available as a Github repository\footnote{https://github.com/yao0510/NLP-2020-EmotionGIF}.
\subsection{Experimental Setup}
For both pre-trained language model and multi-label classification model, we use Adam \citep{kingma2014adam} as optimizer with epsilon 1e-8, learning rate 4e-5.
Gradient accumulation steps and warmup ratio are 1 and 0.06.
Max sequence length, number of epochs, and batch size are set to 113, 4, and 16.

For pre-trained language model, we set block size to 96.
For multi-label classification model, early stopping is used, which means beam search is stopped when number of beam sentences finished per batch.
Early stopping patience is set to 3.
Early stopping metric is eval\_loss and early stopping metric should be minimized.
Most of these configurations are default arguments in Simple Transformers\footnote{https://github.com/ThilinaRajapakse/simpletransformers}.

In order to achieve ensemble methods, we train 10 of \textbf{RoBERTa-base}, 5 of \textbf{BERT-base-cased} and 5 of \textbf{BERT-base-uncased}.
All of these models have been trained with above configurations.
\subsection{Evaluation Metric}
The metric that will be used to evaluate entries is Mean Recall at k, with k=6 (MR@6).
Table \ref{evaluation-example} shows an example how we evaluate our predictions.
For each output, we will predict 6 categories out of a list of 43 categories as \textbf{Prediction} in Table \ref{evaluation-example} and calculate how many categories (N) that our predicted categories are identical to the answer.
The MR@6 is N divided by the total amount of \textbf{Answer}.
The final result is the average of the MR@6 for all Twitter threads.

\begin{table*}
\centering
\begin{tabular}{lll}
\hline \textbf{Answer} & \textbf{Prediction} & \textbf{MR@6} \\ \hline
\textbf{agree}, thank\_you, thumbs\_up & oops, scared, thank\_you, you\_got\_this, do\_not\_want, \textbf{agree} & 1/3\\
\hline
\end{tabular}
\caption{\label{evaluation-example} An example of evaluation metric}
\end{table*}
\begin{table*}
\centering
\begin{tabular}{c|*2c|*2c|*2c}
 & \multicolumn{2}{c|}{\textbf{Train}} & \multicolumn{2}{c|}{\textbf{Validation}} & \multicolumn{2}{c}{\textbf{Test}} \\ 
\hline
Steps & text & reply & text & reply & text & reply  \\
\hline
0 & 77.681\% & 73.164\% & 77.472\% & 72.608\% & 78.101\% & 72.828\% \\
\hline
1 & 77.684\% & 73.166\% & 77.473\% & 72.615\% & 78.102\% & 72.828\% \\
\hline
2 & 80.319\% & 75.973\% & 80.157\% & 75.556\% & 80.592\% & 75.643\% \\
\hline
3 & 92.560\% & 89.777\% & 92.546\% & 89.144\% & 92.731\% & 89.830\% \\
\hline
4 & 93.929\% & 91.692\% & 93.334\% & 91.166\% & 93.455\% & 91.763\% \\
\hline
5 & 93.919\% & 92.086\% & 93.921\% & 91.497\% & 94.056\% & 92.158\% \\
\hline
\end{tabular}
\caption{\label{tokenizer-coverage} Coverage of RoBERTa-base tokenizer}
\end{table*}
\subsection{Preprocessing Analysis}
To check preprocessing methods, we use RoBERTa-base tokenizer coverage to validate shown in Table \ref{tokenizer-coverage}.
From Table \ref{tokenizer-coverage}, training data, validation data, and testing data in EmotionGIF task all have similar coverage in both \texttt{text} and \texttt{reply}.
After applying preprocessing methods, coverage of training \texttt{text} increases from 77.681\% to 93.919\%, \texttt{reply} from 73.164\% to 92.086\%, and coverage of validation \texttt{text} increases from 77.472\% to 93.921\%, \texttt{reply} from 72.608\% to 91.497\%, and coverage of testing \texttt{text} increases from 78.101\% to 94.056\%, \texttt{reply} from 72.828\% to 92.158\%.

In order to further dig into what tokens aren't seen by tokenizer, Table \ref{oov-tokens} shows first 6 out-of-vocab (OOV) tokens.
Although there are still some unknown tokens, we can observe that some words like \texttt{medium-dark} might be able to be tokenized into expected tokens like \texttt{medium}, \texttt{-}, and \texttt{dark}.

Exploratory data analysis (EDA) is a initial investigations on data so as to discover pattern or to check assumption with the help of statistics.
Through EDA, we find that convert word to lower case may cause unexpected tokens from tokenizer.
For example, word \texttt{Hug} can be correctly tokenized when at different positions, while word \texttt{hug} cannot be tokenized as we expected.
That is, \texttt{hug} will be tokenized into \texttt{h} and \texttt{ug}.
Hence we don't convert all words into lower case in EmotionGIF task.

\begin{table}
\centering
\begin{tabular}{ll}
\hline \textbf{Out-of-vocab tokens} & \textbf{Count} \\ 
\hline
pensive & 251 \\
\hline
dependable & 247 \\
\hline
Scotty & 203 \\
\hline
backhand & 194 \\
\hline
6pm & 163 \\
\hline
medium-dark & 152 \\
\hline
\end{tabular}
\caption{\label{oov-tokens} First 6 out-of-vocab tokens}
\end{table}

\subsection{Evaluation Results}
The experiment results of validation data are shown in Table \ref{validation-results}.
The proposed system and baseline are evaluated based on the MAR@6.

To verify our system architecture, we first fine-tune multi-label classification model (MLC) for each type of pre-trained base language model directly.
MLC\textsubscript{RoBERTa}, MLC\textsubscript{BERT-cased}, and MLC\textsubscript{BERT-uncased} all outperform baseline provided by EmotionGIF official.
This explains that directly fine-tune pre-trained model has great performance over baseline.
Also, MLC\textsubscript{RoBERTa} is around 3\% higher than BERT models. 
This shows that RoBERTa does better optimization as compared to BERT.
In order to prevent domain bias, we train tweet dataset on pre-trained language model (LM) to enhance it's knowledge.
Adding language model with RoBERTa (LM\textsubscript{RoBERTa} + MLC\textsubscript{RoBERTa}) improves about 1.2\% on MAR score which implies training tweet data on language model can slightly reach higher performance than directly use formal pre-trained language model.
To reduce affect of dropout weights, we train several models and average them to maintain balance performance.
Motivated from \citet{laurae2020ensemble}, we apply power weighted sum in our system with power 1.8 and weights are 3.0, 1.8, and 0.8 corresponding to RoBERTa, BERT-cased, and BERT-uncased.
With these ensemble methods, our system reaches 0.5619, which improves about 2\% performance on validation set.

Table \ref{testing-results} is our system predict on testing set.
Ensemble models achieve about 0.5662 MAR@6 score, while only using single type of model only gets 0.5404.
This indicates single type of model may be slightly worse in testing data.
Applying ensemble methods does solve this problem.
Overall, our proposed system successfully outperforms with either using original pre-trained language model to fine-tune or EmotionGIF official baseline.
Our approach achieves high MAR@6 score both on validation data and testing data in this competition.

\begin{table}
\centering
\begin{tabular}{ll}
\hline \textbf{Model} & \textbf{MAR@6} \\ 
\hline
Official majority baseline & 0.4009 \\
\hline
MLC\textsubscript{BERT-cased} & 0.5021 \\
\hline
MLC\textsubscript{BERT-uncased} & 0.5023 \\
\hline
MLC\textsubscript{RoBERTa} & 0.5293 \\
\hline
LM\textsubscript{RoBERTa} + MLC\textsubscript{RoBERTa} & 0.5414 \\
\hline
Ensemble models & \textbf{0.5619} \\
\hline
\end{tabular}
\caption{\label{validation-results} Validation results}
\end{table}

\begin{table}
\centering
\begin{tabular}{ll}
\hline \textbf{Model} & \textbf{MAR@6} \\ 
\hline
Official majority baseline & 0.4065 \\
\hline
LM\textsubscript{RoBERTa-base} + MLC\textsubscript{RoBERTa-base} & 0.5404 \\
\hline
Ensemble models & \textbf{0.5662} \\
\hline
\end{tabular}
\caption{\label{testing-results} Testing results}
\end{table}

\section{Conclusion}
In this work, we propose an system architecture combining with preprocessing, model framework, and ensemble models for EmotionGIF task.
We intently convert some words to our desired format and increase the coverage of words recognized by tokenizer.
Based on preprocessing data, We apply multi-label classification and pre-trained model when training models to make our work more sophisticated.
Besides, we also show that ensemble models with power weighted sum outperform any single model with same parameters we trained.

In Section \ref{2}, we observe that there is an imbalance between categories.
However, in the present work, we don't deal with it.
Furthermore, we consider to replace multi-label classification with ranking classification due to its property of dependency in future work.
The probabilities of multi-label classification are treated as independent, so there is no correlation among categories while ranking classification is the opposite.
Since the category would have some connection with each other as Table \ref{correlation table} shown, we assume that it would be better to let our model regard the dependency between categories as critical.

\bibliography{anthology,acl2020}
\bibliographystyle{acl_natbib}
\end{document}